\PassOptionsToPackage{bottom}{footmisc}
\PassOptionsToPackage{dvipsnames}{xcolor}
\documentclass[]{ceurart}

\usepackage{listings}
\usepackage[T1]{fontenc}
\usepackage{graphicx}
\usepackage{lipsum}  
\usepackage{soul}
\usepackage{comment}
\usepackage{caption}
\usepackage{booktabs}
\usepackage{xurl} 
\usepackage{hyperref}
\usepackage[most]{tcolorbox}
\usepackage{amsmath}
\usepackage{amssymb}
\usepackage[group-separator={,},
            group-minimum-digits=4,
            input-decimal-markers={.}]{siunitx}
\usepackage{booktabs}
\usepackage{multirow}
\usepackage[all=normal,bibbreaks=tight,paragraphs=tight,floats=tight,wordspacing=tight,tracking=normal]{savetrees}
\usepackage{array}
\usepackage{adjustbox}
\usepackage{varwidth}
\usepackage{tabularx}
\usepackage[table,dvipsnames]{xcolor}
\usepackage{longtable}
\usepackage{float}
\usepackage[scaled=0.75]{beramono}
\usepackage{placeins}

\newcounter{algorithm}

\begin{document}

\copyrightyear{2022}
\copyrightclause{Copyright for this paper by its authors.
  Use permitted under Creative Commons License Attribution 4.0
  International (CC BY 4.0).}

\conference{20th International Workshop on Ontology Matching, 24th International Semantic Web Conference ISWC-2025 November 2nd or 3rd, 2025, Nara, Japan}

\title{Efficient Uncertainty Estimation for LLM-based Entity Linking in Tabular Data}

\author[1]{Carlo Alberto Bono}[orcid=0000-0002-5734-1274, email=carlo.bono@polimi.it] 
\author[2]{Federico Belotti}[orcid=0009-0008-0140-3318]
\author[2]{Matteo Palmonari}[orcid=0000-0002-1801-5118]

\address[1]{Politecnico di Milano, DEIB, Via Ponzio 34/5, Milano, 20133, Italy}
\address[2]{Università degli Studi di Milano Bicocca, DISCo, Viale Sarca, 336, 20126, Milano}

\begin{abstract}
Linking textual values in tabular data to their corresponding entities in a Knowledge Base is a core task across a variety of data integration and enrichment applications.  Although Large Language Models (LLMs) have shown State-of-The-Art performance in Entity Linking (EL) tasks, their deployment in real-world scenarios requires not only accurate predictions but also reliable uncertainty estimates, which require resource-demanding multi-shot inference, posing serious limits to their actual applicability. As a more efficient alternative, we investigate a self-supervised approach for estimating uncertainty from single-shot LLM outputs using token-level features, reducing the need for multiple generations. Evaluation is performed on an EL task on tabular data across multiple LLMs, showing that the resulting uncertainty estimates are highly effective in detecting low-accuracy outputs. This is achieved at a fraction of the computational cost, ultimately supporting a cost-effective integration of uncertainty measures into LLM-based EL workflows. The method offers a practical way to incorporate uncertainty estimation into EL workflows with limited computational overhead. 
\end{abstract}


\maketitle

\section{Introduction}

Resolving ambiguity in tabular data by linking cell values to entity identifiers in a knowledge base, e.g., a Knowledge Graph (KG) such as Wikidata, is a fundamental integration challenge that impacts many knowledge-driven applications. The task is denoted by different terms:
{\textit{Entity Linking} (EL)} is borrowed from Natural Language Processing (NLP)~\cite{bhagavatula2015tabel,gupta2017entity}, \textit{Entity Reconciliation} designates entity matching across data sources\footnote{\url{https://www.w3.org/groups/cg/reconciliation/}}, and \textit{Cell Entity Annotation} (CEA) concerns the annotation of cell values in tables~\cite{belotti2024evaluating}. The latter is part of the broader objective of understanding tabular data by matching it with a background KG, sometimes referred to as Semantic Table Interpretation (STI)~\cite{liu2023tabular}. 

Large Language Models (LLMs), which have achieved remarkable performance across a variety of NLP tasks, have also been proposed in the context of tabular data understanding~\cite{zhang2024tablellama,li2023table} and EL~\cite{cremaschi2025steellm,jayawardene2024tablinkllm,nararatwong2024evaluating,belotti2024evaluating}, where an LLM is asked to link values to the correct entity from a set of candidates retrieved by dedicated components. Evidence suggests that medium-sized LLMs fine-tuned on the task and general-purpose top-tier LLMs achieve state-of-the-art results and generalization capabilities on different benchmarks~\cite{belotti2024evaluating}. 

However, LLM-based approaches to EL usually return only the label of the selected candidate, without providing explicit insights into the uncertainty associated with the model's decision. This limitation is exacerbated by the non-determinism in the generation process, since multiple runs on the same input may return different answers. 
In EL and similar matching and classification tasks, previous approaches for uncertainty estimation, including those using language models in combination with classifiers, e.g., TURL~\cite{deng2022turl}, associate the link with some confidence score. Confidence scores are explicitly included in the Reconciliation API\footnote{\scriptsize\url{https://www.w3.org/community/reports/reconciliation/CG-FINAL-specs-0.2-20230410/}} proposed by the W3C Entity Reconciliation Community Group. Since a human-in-the-loop process is frequently used to improve the quality of links in quality-critical applications~\cite{klie2020zero,arnold2016tasty}, the presence of confidence scores enables the identification of potential errors by prioritizing links that humans could review to improve the quality of the results. 
Understanding and quantifying the uncertainty of the LLM output is
essential to ensure the robustness and trustworthiness of the model output, to highlight where the model is fragile or ungrounded, and to direct human intervention in order to maximize its efficacy. 

An established way of measuring uncertainty in LLMs is to determine the likelihood of an answer being consistent over multiple independent generations~\cite{manakul2023selfcheckgpt,farquhar2024semantic,huang2024survey}. In the context of LLMs, this introduces significant computational overhead, as the generation time scales linearly with the number of tokens in both the prompt and the output, even with caching techniques~\cite{luohe2024keep}. 
Additionally, if we consider that EL may be applied to large datasets, this approach may be unsustainable. To mitigate the resource demands of large-scale EL scenarios, we propose an efficient approach to estimate uncertainty in LLMs without relying on multiple generations during inference. More precisely, we propose an efficient self-supervised method that learns to estimate the uncertainty observed over multiple generations of an LLM using observables from a single generation. The method leverages token-level features, derived from the probability distribution over the output vocabulary, to train a lightweight regression model targeting the ``true'' observed uncertainty.  

Although in this paper we focus on EL on tabular data, we believe that our method could be applied to different closed-form tasks that can benefit from uncertainty quantification.

\textcolor{black}{This paper makes the following contributions:
\begin{itemize}
  \item Formalizes uncertainty-aware EL on tabular data and proposes a self-supervised regressor that learns to approximate multi-shot uncertainty from single-shot token-level features;
  \item Introduces a lightweight, model-agnostic feature set from output-layer probabilities and optional intermediate-layer signals, requiring no task labels;
  \item Evaluates the approach across several instruction-tuned LLMs, showing how uncertainty-awareness can effectively lead to the detection of low-accuracy outputs, which can then be corrected to improve accuracy under a constrained review budget.
\end{itemize}}

\section{Related Work}
\label{sec:related}

\paragraph{Uncertainty in EL for Tabular Data.} Associating confidence scores in matching tasks, including EL for tabular data, is a well-established practice~\cite{batini2016data,li2017human,liu2023tabular}, where scoring a list of retrieved candidates is a typical intermediate step. 
This is also true in approaches combining a pretrained language model with classifiers, such as TURL~\cite{deng2022turl} and UNICORN~\cite{fan2024unicorn}.
While scores computed in pre-LLMs methods somehow support confidence estimation, less attention has been dedicated to systematic analyses of uncertainty estimation in this task. One approach based on a deep neural network proposes a supervised method that considers the matching score of the best candidate and its distance from the second-best to quantify confidence~\cite{avogadro2023estimating}. The authors show that computed confidence scores help prioritize links to revise with a progressive budget. To the best of our knowledge, recent approaches based on LLMs~\cite{cremaschi2025steellm,jayawardene2024tablinkllm,nararatwong2024evaluating,belotti2024evaluating,zhang2024tablellama} have not investigated how to exploit uncertainty measures in generative approaches. Uncertainty measures may support the decision whether to link or not a top candidate (e.g., based on a threshold), including the detection of NIL entities (i.e., values associated with entities not in the KB) ~\cite{liu2023tabular}. Additionally, little attention has been dedicated to the impact of the variability of the links predicted by LLMs under different generations within this task.


\paragraph{Uncertainty Estimation for LLMs.} \textcolor{black}{\textit{Confidence Score} methods leverage single‑shot proxies from output probabilities (e.g., entropy, log‑probabilities, perplexity) but can be overconfident when wrong~\cite{huang2025leap,manakul2023selfcheckgpt,lin2023generating,ma2025logits,plaut2025miscalibrated}; \textit{Semantic Consistency} approaches, involving the generation of multiple outputs for the same prompt and measuring their consistency or semantic similarity, are effective but computationally expensive~\cite{lin2023generating,farquhar2024semantic}; \textit{Supervised} methods learn calibrated uncertainty from features of generated text or hidden states and often outperform unsupervised heuristics, but require an annotated dataset~\cite{vazhentsev2025token,liu2024uncertainty}; \textit{Ensemble‑ and Bayesian‑Inspired} approaches estimate approximate Bayesian uncertainty (e.g., deep ensembles, Monte-Carlo dropout) but are generally impractical at LLM scale~\cite{fadeeva2023lm,lakshminarayanan2017simple,srivastava2014dropout}; \textit{Verbalized and Self‑Reported Uncertainty} methods improve interpretability by prompting models to report confidence, with mixed reliability across tasks~\cite{kadavath2022mostly,lin2022teaching,xiong2023can}. Finally, ~\cite{yang2023improving,yadkori2024believe} employ \textit{Uncertainty‑Aware In‑Context Learning} to filter/refine or to guide iterative prompting, improving reliability on open‑ended tasks}.

\paragraph{This work.} Our approach intersects multiple-generations and supervised paradigms by leveraging a self-supervised uncertainty \textcolor{black}{regressor} that learns from multiple-generations outputs how to estimate uncertainty from a single generation. Although this work shares some conceptual similarities with~\cite{liu2024uncertainty}, our method does not require supervised labels and focuses on tabular EL.

\section{Methods}
\label{sec:methods}


\subsection{Problem Formulation}
\label{sec:problem}

\textcolor{black}{Let a tabular EL instance be defined by an input tuple $(T, m, E_{\text{KG}})$, where $T$ is a table, $m$ is a mention at coordinates $(r,c)$, and $E_{\text{KG}}=\{e_1,\dots,e_K\}$ is a retriever-provided set of candidate entities, dependent on the particular Knowledge Graph $\text{KG}$. An LLM $M$ conditions on a prompt $\mathbf x$ built from $(T, m, E_{\text{KG}})$ and generates a textual answer $y$ from which a deterministic post-processor extracts a selected candidate $\hat{e}(y) \in E_{\text{KG}}$. For non-deterministic decoding, repeated generations $\{y^{(i)}\}_{i=1}^N$ induce an empirical distribution $\hat{p}_\text{ans}(\cdot | \mathbf{x})$ over answers\footnote{\textcolor{black}{In particular, ${\hat{p}_\text{ans}(y| \mathbf x)=\frac{1}{N}\sum_{i=1}^N \mathbb{I}[y^{(i)}=y]}$, where $\mathbb{I}[x]=1$ if the predicate $x$ holds, and $0$ otherwise.
}}, from which an uncertainty $u(\cdot)$ can be estimated. Here, \(u(\cdot)\) is the uncertainty score derived from the empirical distribution of generations\footnote{In Section~\ref{sec:target} we instantiate it using the entropy of \(\hat p_{\text{ans}}\) and, via semantic grouping~\cite{farquhar2024semantic}, of \(\hat p_{\text{sem}}\).}. Our goal is to output, for each $\mathbf x$, both a selected candidate $\hat{e}(y)$ and an uncertainty score $\hat{s}(y)$ that correlates with $u(\cdot)$ computed a posteriori from multiple generations. We learn this score via an uncertainty regressor $h_\phi$ from single-shot token-level features (cf. Section~\ref{sec:observables}). At deployment, we use $\hat{s}(y)$ to flag instances for manual review. Algorithm~\ref{algo:ssl-uc} outlines the complete training (warm-up) and inference workflow.}

\subsection{Measures of Uncertainty} 
\label{sec:target}

To quantify \textcolor{black}{the uncertainty $u(\cdot)$} of LLM outputs, we adopt two widely recognized measures: Predictive Entropy (PE), which captures the uncertainty inherent to a model’s answer distribution, and Semantic Entropy (SE), which considers semantic equivalence classes rather than raw answers for measuring uncertainty \cite{farquhar2024semantic}. PE is defined as the entropy of the observed output sequences $y$, realizations of a random variable $Y$, conditional on an input sequence $\mathbf{x}$, such that:
\begin{equation}
\label{eq:pe}
PE(\mathbf{x}) = \textcolor{black}{H(\hat{p}_\text{ans}(\cdot|\mathbf{x})) = - \sum_{y}\hat{p}_\text{ans}(y|\mathbf{x})\log \hat{p}_\text{ans}(y|\mathbf{x})}
\end{equation}
Lower PE indicates that the output distribution is concentrated around a single answer, while higher PE reflects uncertainty over multiple possible answers. SE is instead calculated over the distribution of answer meanings $\hat{p}_\text{sem}(\cdot | \mathbf{x})$\footnote{Let $g:\mathcal{Y} \to \mathcal{C}$ map an answer $y$ to its semantic class $c=g(y)$~\cite{farquhar2024semantic}. The empirical distribution over semantic classes is
${\hat{p}_{\text{sem}}(c | \mathbf{x}) = \sum_{y \in \mathcal{S}(\mathbf{x})} \hat{p}_\text{ans}(y|\mathbf{x})\mathbb{I}[g(y)=c] = \frac{1}{N}\sum_{i=1}^N \mathbb{I}[g(y^{(i)})=c]}$, where $\mathcal{S}(\mathbf{x})$ is the set of observed distinct answers.}, that is, over the semantic equivalence classes $c$ observed on the output, where each class refers to a shared meaning:
\begin{equation}
\label{eq:se}
SE(\mathbf{x}) = \textcolor{black}{H(\hat{p}_\text{sem}(\cdot|\mathbf{x})) = - \sum_{c}\hat{p}_\text{sem}(c|\mathbf{x})\log \hat{p}_\text{sem}(c|\mathbf{x})}
\end{equation}
where $SE(\mathbf x)$ is the same as in \cite{farquhar2024semantic}. To enable a comparison between entropy measures computed over discrete distributions with varying support sizes, both measures are normalized by the logarithm of the number of unique outputs.
We also report the sequence perplexity (PP) as a confidence baseline. For an answer
\(y=(y_1,\dots,y_T)\) of $T$ tokens,
\begin{equation}
PP(y)
=\exp\!\left(-\frac{1}{T}\sum_{t=1}^{T}\log p\big(y_t | y_{<t}, \mathbf{x}\big)\right)
\end{equation}
where \(p(y_t | y_{<t},\mathbf{x})\) is the model’s next-token probability.


\subsection{Uncertainty Estimation}
\label{sec:estimationmethod}

\begin{figure}[!t]
    \centering
    \includegraphics[width=0.45\textwidth]{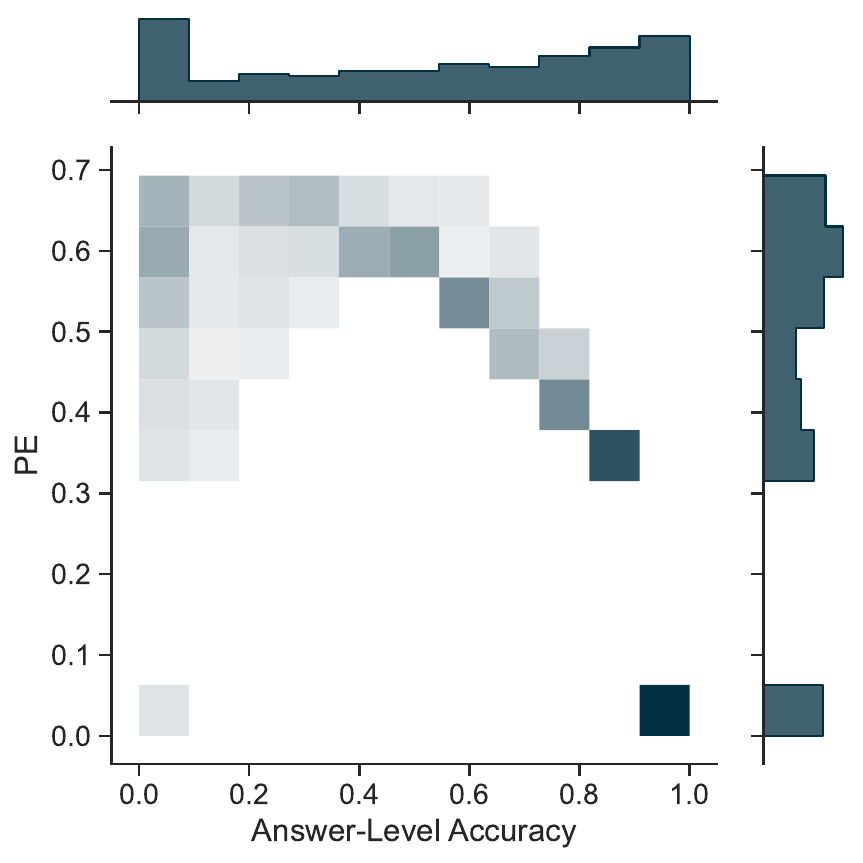}
    \includegraphics[width=0.45\textwidth]{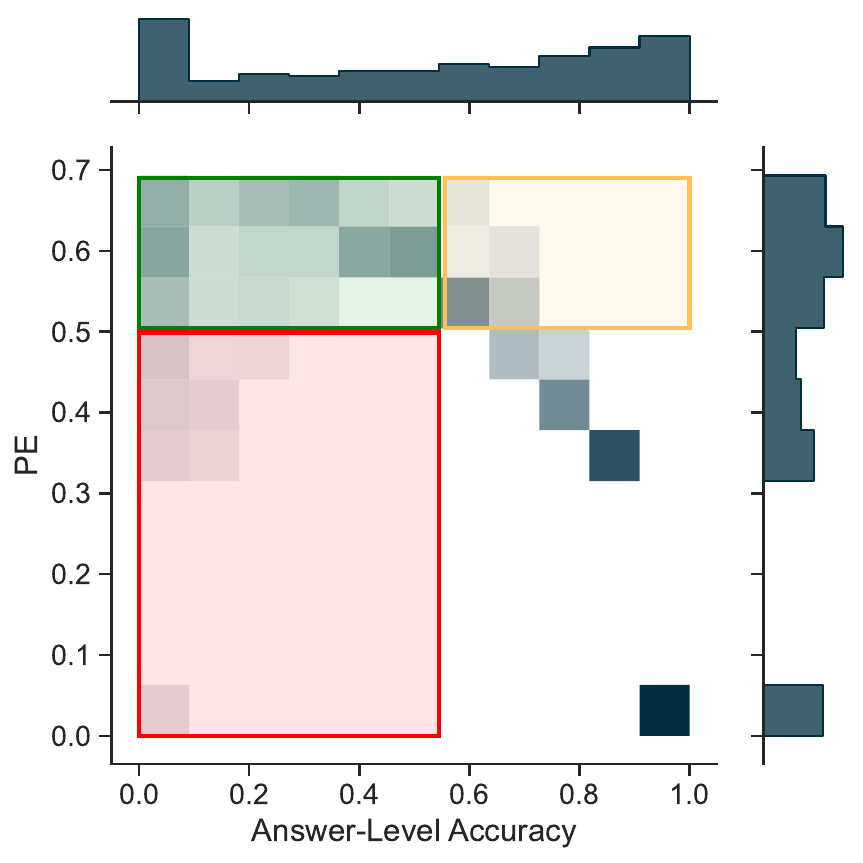}
    \caption{\textcolor{black}{\textbf{(left)} Joint distribution of answer-level accuracy and Predictive Entropy (PE) per input for \textit{Llama-3.1-8B-Instruct}, with marginal distributions on the axes. \textbf{(right)} We define the positive class as “flag for manual review”. Thresholding uncertainty yields: green = correctly flagged low-accuracy cases (true positives), yellow = incorrectly flagged high-accuracy cases (false positives), red = consistently wrong but low-uncertainty cases (false negatives; not recoverable by uncertainty thresholding).}}
    \label{fig:accuracy_vs_variability}
\end{figure}

Our practical objective is to estimate uncertainty and use it to flag uncertain answers for manual review, without relying on an explicit ground truth. The underlying intuition is that higher uncertainty is correlated with lower accuracy, while low uncertainty alone does not imply higher accuracy. Accuracy is calculated as the average answer correctness across $N$ generated outputs for each input prompt. 

\paragraph{Uncertainty–Accuracy Relationship.}\textcolor{black}{ Figure~\ref{fig:accuracy_vs_variability} highlights the effect of setting a threshold on uncertainty to flag items for manual review. Interpreting the flag decision as a binary classifier, the \textbf{green} area contains correctly flagged low-accuracy cases (true positives). Reviewing these comes at the cost of also reviewing cases in the \textbf{yellow} area (false positives), where accuracy would be satisfactory but uncertainty is high. 
Cases in the \textbf{red} area are consistently wrong but with low variability; these are false negatives and are not recoverable by uncertainty-based thresholding. Intuitively, lowering the uncertainty threshold expands the region of flagged cases. On the other hand, raising the threshold shrinks both green and yellow zones, reducing workload at the risk of missing additional bad cases. Notably, the subset of red points with zero uncertainty, which are consistently wrong across generations, remains invariant to the threshold and unrecoverable by uncertainty thresholding alone. Sweeping the threshold traces the ROC analysis discussed later.}

\paragraph{Efficient Uncertainty Estimation via Self‑Supervised Regression.} A large class of uncertainty estimation methods, including widely used metrics like PE and SE, depend on repeated sampling for their computation, which comes with significant resource overhead. We therefore propose an efficient self-supervised method to approximate such uncertainty measures using information from a single generation, making them more suitable for downstream use. The target variable -- either PE or SE, measured on $N$ independent runs of an LLM -- is regressed from observable features using a Random Forest model implemented via \texttt{XGBRFRegressor}, with 100 estimators \cite{breiman2001random,xgboost}. \textcolor{black}{We optimize the Mean Squared Error (MSE) between the predicted score $\hat{s}(y)$ and the normalized multi-shot target $u(\hat{p}_\text{ans})$ (Section~\ref{sec:target})}. We utilize 10-fold cross-validation grouped by prompt to avoid data leakage and to obtain reliable performance estimates on unseen cases. The \textcolor{black}{regressor} is trained in a warm-up phase, during which $N$ generations per prompt are collected and the derived target variable is learned. At runtime, the regressor estimates the uncertainty from a single generation in response to the same prompt. \textcolor{black}{We denote the learned uncertainty regressor by \(h_\phi\), which maps features \(F(M,\mathbf x)\) extracted from a target model \(M\)’s single generation \(y=M(\mathbf x)\) to an estimate \(\hat s(y)=h_\phi(F(M,\mathbf x))\).} The method is self-supervised as it does not require externally labeled data. Moreover, the regression incurs negligible computational overhead. Further considerations regarding learning convergence are provided in Section~\ref{sec:learnability}.

We collect the features, as defined in Section~\ref{sec:observables}, from $N$ independent runs of an LLM. We benchmark the proposed method considering: (1) the first 10 generated tokens, and (2) all the tokens from the fixed-width tail of the prompt\footnote{The token indices collected in (2) are fixed across models, whereas those in (1) are model-dependent.}. This allows us to investigate the contribution of different prompt segments to the quality of the uncertainty estimates. We also evaluate different feature sets as input, as detailed in Section~\ref{sec:settings}, to assess the relative importance of individual feature groups.

\subsection{\textcolor{black}{Features for Uncertainty Estimation}}
\label{sec:observables}

We rely on \textcolor{black}{features $F(M, \mathbf x)$} that are observable or easily computable during the inference process of an LLM $M$. We compute token-wise features for specific portions of the prompt and generated tokens. Instead of aggregating the features at the sequence level \cite{manakul2023selfcheckgpt}, we maintain the features at the token level. 
Let \(V\) be a vocabulary of tokens and let \(\mathbf{x}\) be a sequence of tokens from \(V\). Given the input prompt \(\mathbf{x}\), an LLM computes a logit vector \(\mathbf{z}=M(\mathbf{x}) \in \mathbb{R}^{|V|}\) for each token. It then applies a softmax function to derive the stochastic vector $\mathbf{p}$, i.e., the probability of \(t\) being the next token for every $t \in V$:
\[
\mathbf{p} \in \mathbb{R}^{|V|} \text{ s.t. } p_t = \sigma(\mathbf{z})_t = \frac{\exp(z_t/{\tau})}{\sum_{v\in[|V|]}\exp(z_v/{\tau})}
\]
where $\tau$ is the temperature parameter, which controls the randomness of the output distribution. We calculate for each token\footnote{We use the notation $[K]$ to indicate $\{1,..., K\}$.}:
\begin{itemize}
        \item \textit{Max probability}. The maximum probability observed across the output vocabulary, measured at the output layer $L$ of the model.

        \begin{equation*}
            \text{M}(\mathbf{p}) = \max_{t \in [|V|]}\ p_t
        \end{equation*}

        \item \textit{Entropy}. The entropy of the token probability distribution

        \begin{equation*}
            \text{H}(\mathbf{p}) = - \sum_{t \in [|V|]} p_t\log p_t
        \end{equation*}

        over all vocabulary tokens at the output layer $L$. Lower entropy indicates higher confidence, while higher entropy reflects more uncertainty.
        
        \item \textit{LogitLens}~\cite{nostalgebraist2020logitlens}. The Kullback-Leibler (KL) divergence between the probability distribution at each intermediate layer $l$ and the probability distribution at the output layer $L$ \cite{nostalgebraist2020logitlens}.

        \begin{equation*}
        \text{KL}(\mathbf{p}^l \parallel \mathbf{p}^L) = \sum_{t \in [|V|]} p_t^{l}  \log \frac{ p_t^{l} }{ p_t^{L} }
        \end{equation*}

        where $\mathbf{p}^{l}$ is measured for each layer $l \in [L-1]$ and $\mathbf{p}^{L}$ is measured on $L$, resulting in $L-1$ divergence measures quantifying the deviation between the distribution at the intermediate layers and the output layer. At each layer $l$, $\mathbf{p}^{l} = \sigma(\mathbf{z}^l)=\sigma\left(\text{LayerNorm}[\mathbf{h}^l]W_U\right)$ is obtained by applying the last LayerNorm to the hidden representation $\mathbf{h}^l$ and multiplying it by the unembedding matrix $W_U$.
\end{itemize}

\refstepcounter{algorithm}\label{algo:ssl-uc}
\begin{tcolorbox}[floatplacement=ht,float,title=Workflow overview for self-supervised uncertainty estimation,
colback=gray!5!white,colframe=gray!80!black,fonttitle=\bfseries]
\textbf{Warm-up (offline)}
\begin{enumerate}
  \item \textcolor{black}{For each prompt $\mathbf x$, collect $N$ independent generations from the LLM $M$ to obtain $\{y^{(i)}\}_{i=1}^N$.}
  \item \textcolor{black}{Compute the a posteriori uncertainty target $u(\hat{p}_\text{ans/sem})$ either as normalized PE or SE (Equations~\ref{eq:pe}~and~\ref{eq:se}, respectively).}
  \item \textcolor{black}{For each generation $i\in[ N ]$, extract per-generation features $F^{(i)}(M, \mathbf x)$ using token-level observables (Section~\ref{sec:observables}) over selected segments (e.g., \textit{Postilla}, first $G$ \textit{Generated} tokens).}
  \item \textcolor{black}{Form training pairs $\big(F^{(i)}(M, \mathbf x),\,u(\hat{p}_\text{ans/sem})\big)$ for $i=1,\dots,N$ and for all prompts $\mathbf x$; train the regressor $h_\phi$ on the training pairs.}
\end{enumerate}
\textbf{Inference (online)}
\begin{enumerate}
  \item \textcolor{black}{For each new prompt $\mathbf x$, run a single generation to produce an answer $y=M(\mathbf x)$ and extract $F(M, \mathbf x)$.}
  \item \textcolor{black}{Return $y$ and predict $\hat{s}(y)=h_\phi(F(M, \mathbf x))$ and use it to decide if the item should be reviewed.}
\end{enumerate}
\end{tcolorbox}

\subsection{Runtime Considerations}
\label{sec:runtime}

\textcolor{black}{At inference time, the proposed approach reduces the number of LLM generations used to estimate uncertainty from $N$ to 1 and adds only a lightweight regression pass on features extracted during generation. In typical settings, the regression overhead is negligible compared to a single forward pass of the LLM; thus, single-shot estimated PE/SE achieves most of the benefit of multi-shot uncertainty at substantially lower cost. For a theoretical analysis of Transformer time complexity with and without KV-cache during generation, see Appendix~\ref{app:complexity}.}

\section{Experimental Evaluation}
\label{sec:results}

\textcolor{black}{In this work, we address the following research questions:}
\begin{itemize}
  \item[\textbf{Q1}] \textcolor{black}{Do our single-shot uncertainty estimates identify low-accuracy answers?}
  \item[\textbf{Q2}] \textcolor{black}{How much does uncertainty-guided manual correction improve accuracy under a budget $B$?}
  \item[\textbf{Q3}] \textcolor{black}{How much warm-up data is needed to learn the uncertainty regressor $h_\phi$ effectively?}
  \item[\textbf{Q4}] \textcolor{black}{How does temperature $\tau$ affect the uncertainty/accuracy trade-off?}
\end{itemize}
\textcolor{black}{Reproducible code and plotting scripts are available at: \url{https://github.com/carloalbertobono/llm-uncertainty}}.

\subsection{Dataset}
\label{sec:dataset}

We address the Entity Linking (EL) task on tabular data, \textcolor{black}{formalized in Section~\ref{sec:problem}}, where each table mention must be linked to its corresponding entity in a Knowledge Base. The dataset used, \textit{TableInstruct-EL-2K}, is adapted from the TableInstruct EL test set~\cite{zhang2024tablellama}, which is formatted for LLMs and contains 2{,}000 mentions annotated with Wikidata entities, with exactly one correct entity per mention. The original dataset included $\sim$600 mentions with only one candidate (i.e., the correct one), which limits our ability to evaluate answer variability. To address this issue, the candidates were enriched with those retrieved using LamAPI~\cite{avogadro2022lamapi}, a full-fledged retriever that returns richer candidate sets. In the resulting dataset, \num{1650} mentions ($\sim$91\%) include at least 45 candidate entities each. Prompts follow the ``Entity Linking'' template from~\cite{zhang2024tablellama} and include the following segments: an \textit{Instruction} that provides context and task guidelines; an \textit{Input} table, Markdown-serialized; a \textit{Question} asking which of the provided referent candidates corresponds to a specific table mention; a fixed-width \textit{Postilla} that clarifies the expected answer format. The \textit{Generated} segment contains the model’s predicted answer. \textcolor{black}{To support understanding, a minimal, self-contained illustrative example of the prompt, candidate list, and expected answer format is provided in Appendix~\ref{sec:illustrative}.}

\subsection{Experimental Settings}
\label{sec:settings}

Experiments are performed using the following models: \textit{Gemma-2-2B-Instruct}, \textit{Gemma-2-9B-Instruct}, \textit{Llama-3.1-8B-Instruct}, \textit{TableLlama}, \textit{Qwen2.5-7B-Instruct}. These models were chosen as a representative set of instruction-tuned, open-source language models. Additionally, \textit{TableLlama} is considered state-of-the-art in the Entity Linking task on tabular data~\cite{belotti2024evaluating}. The number of generations for each prompt is set to $N=10$, while the temperature $\tau$ is set to $1.0$. Features (observables) are extracted over different portions of tokens, specifically from the \textit{Postilla} and from the first $10$ tokens of the \textit{Generated} segment\footnote{The performance differs depending on whether features are extracted from the \textit{Postilla} or \textit{Generated} tokens. A more detailed analysis can be found in Appendix~\ref{app:position-dep-contrib}.}, considering features from the output layer alone (M$(\mathbf{p})$ and H$(\mathbf{p})$), from the intermediate layers alone (LogitLens), and combined. These combinations are motivated by the hypothesis that uncertainty may manifest differently across prompt segments and feature groups, necessitating an empirical assessment of their impact. We compare the entropy values derived with our method against the following baselines: PE, SE, PP, and an oracle with access to the true answers.

In the remainder of the paper, we utilize two different concepts of accuracy. \textbf{Answer-level} accuracy refers to the accuracy of the answers to a given prompt, computed over multiple generations. Answer-level accuracy can then be aggregated at the level of a set of prompts; we refer to \textbf{dataset-level accuracy} to indicate the average answer-level accuracy computed over the whole dataset.

\begin{table}[!t]
\centering
\small 
\setlength{\tabcolsep}{4pt} 
\caption{\textcolor{black}{Distribution of answer-level accuracy over $N{=}10$ non-deterministic runs, grouped by recoverability. \textit{Unrecoverable} cases (``Always correct'' and ``Never correct (w/o Unc.)'') have zero observed output variability; \textit{Recoverable} cases (``Never correct (w/ Unc.)'' and ``Sometimes correct'') exhibit variability that uncertainty thresholding can surface. The rightmost column reports the review-eligible fraction (Recoverable total $=$ \textit{Never correct (w/ Unc.)} $+$ \textit{Sometimes correct}). Rows sum to 1.}}
\label{tab:accuracy_distribution}
\begin{tabularx}{\linewidth}{l *{5}{>{\centering\arraybackslash}X}}
\toprule
 & \multicolumn{2}{c}{\textbf{Unrecoverable}} & \multicolumn{2}{c}{\textbf{Recoverable}} & \\
\cmidrule(lr){2-3} \cmidrule(lr){4-5}
\textbf{Model} & \textit{Always correct} & \textit{Never correct (w/o Unc.)} & \textit{Never correct (w/ Unc.)} & \textit{Sometimes correct} & \textbf{Recoverable total} \\
\midrule
Gemma-2-2B-Instruct & 0.18 & 0.06 & 0.36 & 0.40 & 0.76 \\
Gemma-2-9B-Instruct & 0.67 & 0.10 & 0.09 & 0.14 & 0.23 \\
Llama-3.1-8B-Instruct & 0.15 & 0.01 & 0.17 & 0.67 & 0.84 \\
Qwen2.5-7B-Instruct & 0.40 & 0.02 & 0.15 & 0.43 & 0.58 \\
TableLlama & 0.79 & 0.03 & 0.01 & 0.17 & 0.18 \\
\bottomrule
\end{tabularx}
\end{table}

\subsection{Task Accuracy and Recoverable Errors}
\label{sec:accuracy}

Regarding the performance of the different LLMs on the task, Table~\ref{tab:accuracy_distribution} summarizes the proportion of answers with and without uncertainty -- \textit{Recoverable} and \textit{Unrecoverable}, respectively -- considering ${N=10}$ runs per item. Answers with no associated uncertainty are grouped into \textit{Always correct} and \textit{Never correct}; answers with associated uncertainty are grouped into \textit{Never correct} and \textit{Sometimes correct}. \textcolor{black}{The rightmost column (Recoverable total) adds \textit{Never correct (w/ Unc.)} and \textit{Sometimes correct} to give the recoverable fraction, i.e., the share of items that uncertainty thresholding can, in principle, surface for manual correction.} It should be noted that \textit{Never correct} items are partitioned into zero-variance items (\textit{w/o Unc.}) and items with observed uncertainty (\textit{w/ Unc.}). 

\textit{Gemma-2-2B-Instruct}, the smallest model in terms of parameters, and \textit{Llama-3.1-8B-Instruct} show the lowest \textit{Always correct} proportion, suggesting limited reliability. In contrast, \textit{TableLlama} and \textit{Gemma-2-9B-Instruct} show the highest \textit{Always correct} proportion, with \textit{Qwen2.5-7B-Instruct} falling in between. \textcolor{black}{The Recoverable total highlights the best-case pay-off of uncertainty-guided review.} In general, the fraction of unrecoverable-never-correct cases remains modest.

\subsection{\textcolor{black}{Q1 and Q2: Uncertainty Estimates Assessment}}
\label{sec:main}

We assess the reliability of our estimates through a series of targeted experiments. First, we test whether the uncertainty measures can identify mentions with low answer-level accuracy, where ``low'' means that the answer-level accuracy is below 0.5 over $N=10$ generations. To this end, we perform a ROC analysis~\cite{hajian2013receiver} on both the estimated and baseline entropy measures. We report the results in terms of true positive rate (TPR) and false positive rate (FPR), where TPR corresponds to the actual low-accuracy cases that the uncertainty-based method correctly flags. At the same time, FPR is the fraction of high-accuracy cases (accuracy $\ge 0.5$) that the method wrongly flags as low-accuracy. Here, the positive class corresponds to the decision “flag for manual review.” In the ROC plot, each curve traces the trade-off between TPR and FPR as a threshold on the uncertainty score—used to decide “flag or not”—is varied. The diagonal line represents random chance; curves that bow upward (high TPR at low FPR) indicate effective uncertainty signals for identifying low-accuracy items.

Moreover, we assess the impact of the uncertainty measures on budget-dependent manual correction. To this end, we investigate the average accuracy obtained by selecting the most uncertain items up to a budget $B$, according to a given uncertainty measure. We assess the aggregate accuracy at the dataset level after correcting a fraction $B$ of the items.

\paragraph{\textcolor{black}{Detecting Low-Accuracy Answers}}
\begin{figure}[!t]
    \centering
    \includegraphics[width=0.9\textwidth]{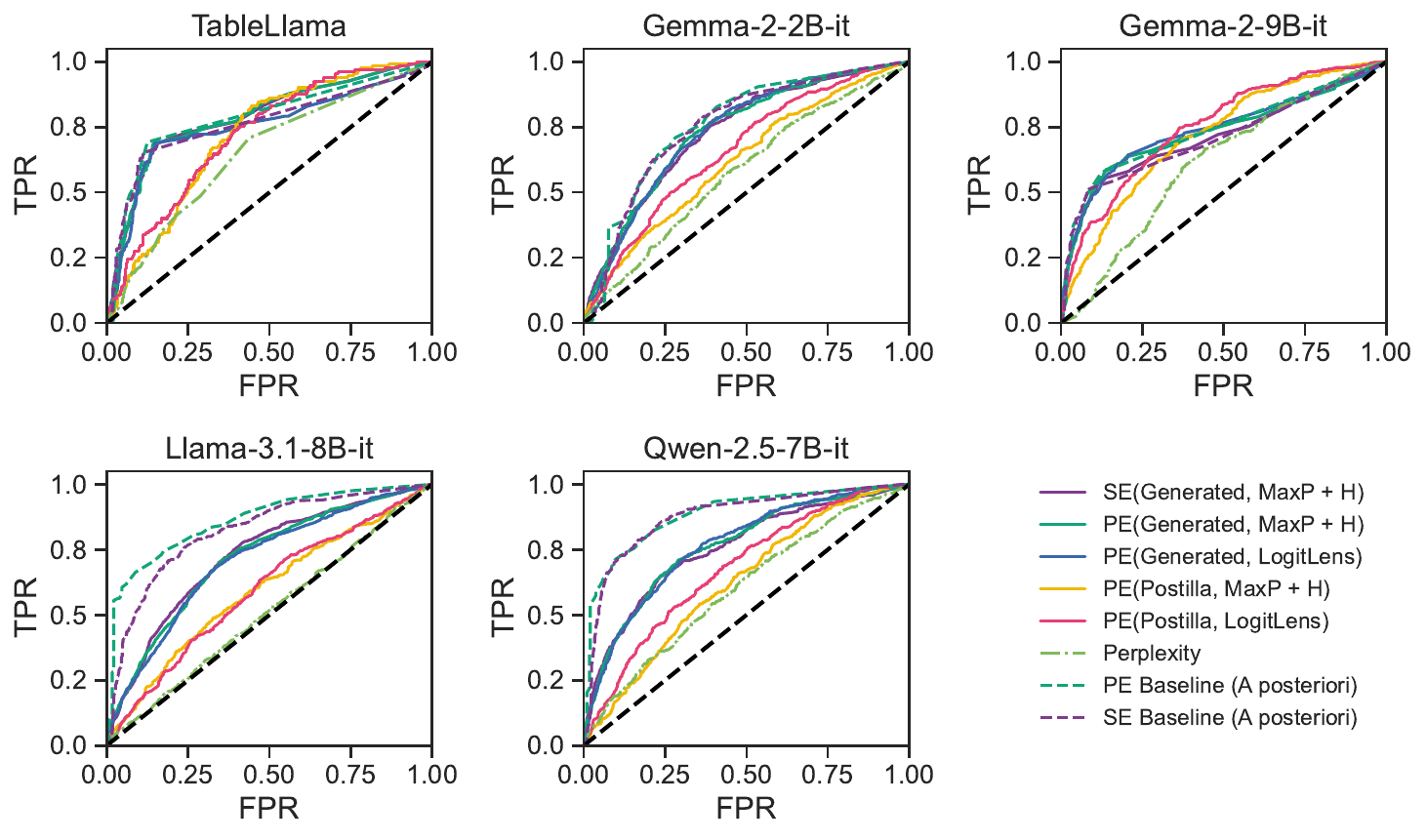}
    \caption{\textcolor{black}{ROC analysis targeting low-accuracy ($<0.5$) cases for selected models. Positive = “flag for manual review”. In the legend, the notation ``\texttt{Target(Segment, Observable)}'' indicates that the target variable \textit{Target} was predicted using a regressor trained on \textit{Observable} features extracted from the \textit{Segment} portion of the prompt. Dashed lines referring to \textit{PE/SE Baseline (a posteriori)} represent the multiple-generations PE and SE computed over $N=10$ generations.}}
    \label{fig:performance}
\end{figure}
\begin{figure}[!t]
    \captionsetup{font={small}}
    \centering
    \includegraphics[width=0.9\textwidth]{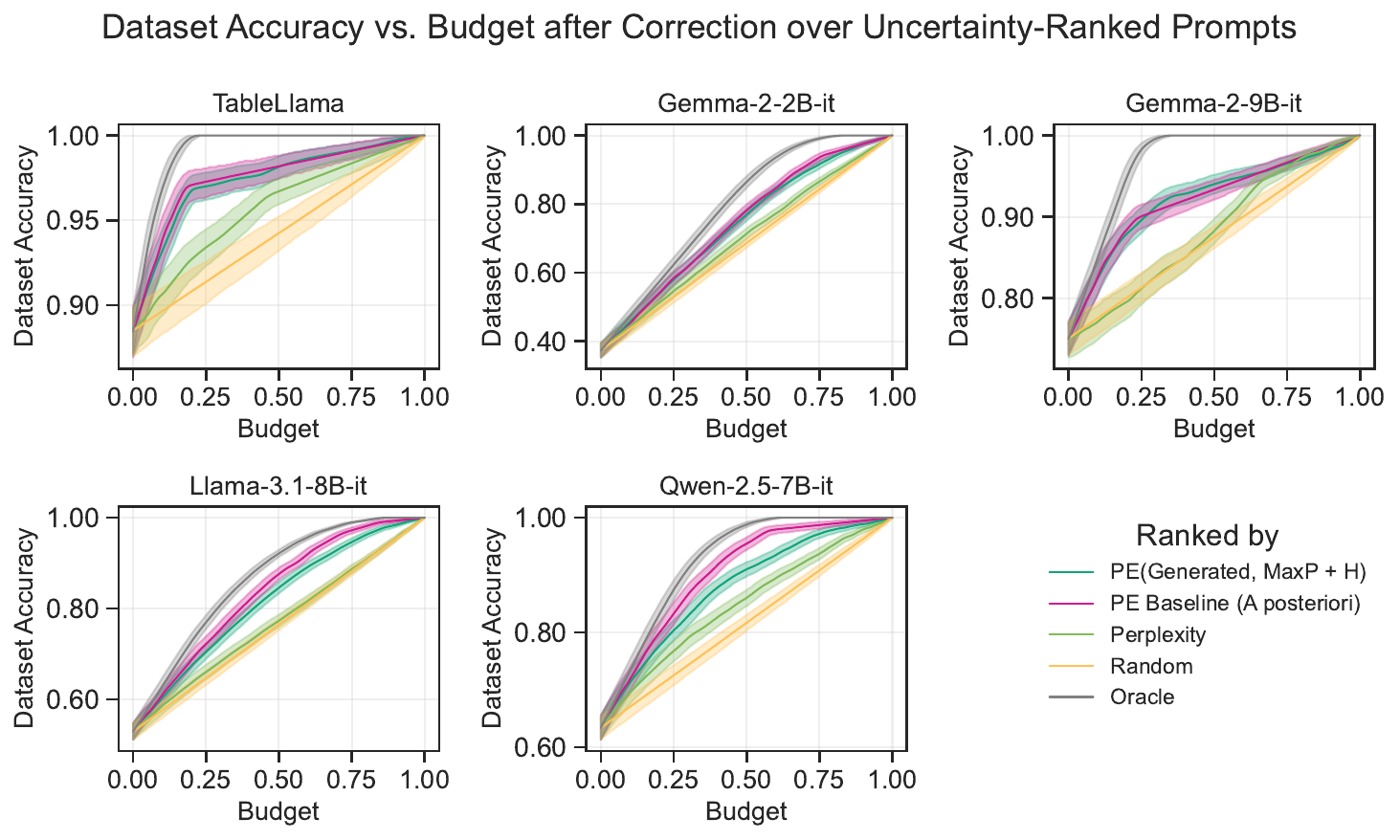}
    \caption{Dataset accuracy as a function of the budget $B$ of items corrected, where ranking is based on the measures shown in the legend. The shaded areas correspond to the 95\% C.I. estimated via \num{1000} bootstrap resampling iterations~\cite{keeping1995introduction}. Each curve illustrates how accuracy improves as more high-uncertainty prompts are corrected. In the legend, the notation ``\texttt{Target(Segment, Observable)}'' indicates that the target variable \textit{Target} was predicted using a regressor trained on \textit{Observable} features extracted from the \textit{Segment} portion of the prompt. \textit{PE/SE Baseline (a posteriori)} represent the multiple-generations PE and SE computed over $N=10$ generations.}
    \label{fig:performance_corrected}
\end{figure}

We evaluate how well the uncertainty estimate can identify \textit{low accuracy} items (answer-level accuracy $\leq 0.5$). Figure~\ref{fig:performance} shows the ROC curves for the considered LLMs, comparing the different uncertainty estimates and baselines. ROC curves are a well-established metric for assessing classification performance, and illustrate the trade-off between true positive rate and false positive rate across different thresholds, in our case, on uncertainty estimates. 
PE and SE based on $N$ generations (dashed lines) show the highest performance. The gap between these baselines and the estimated PE/SE (solid lines) captures the performance loss due to estimating the multiple-generations entropy with the information from a single generation. Notably, when using the best-performing models, \textit{TableLlama} and \textit{Gemma-2-9B-Instruct}, PE and SE computed on the $N$ multiple generations closely approximate the estimated PE and SE (cf. green and purple solid and dashed lines). However, across other models, the PE and SE baselines show a better ability to distinguish high- and low-accuracy cases. Moreover, PP is consistently outperformed by other methods. In general, the performance of our method falls between PE/SE and PP. We recall that the estimates of our method are obtained at a fraction of the computational cost, by computing $N=1$ generations instead of $N=10$.

Regarding the use of features derived from different stages of the generation process, estimates based on the features observed in the \textit{Postilla} segment appear to be generally less informative compared to the \textit{Generated} segment. However, focusing on the utilization of the LogitLens features, we highlight that their contribution is substantial when using the \textit{Postilla} features, while it becomes negligible when using the \textit{Generated} features. 

\paragraph{\textcolor{black}{Uncertainty-Guided Correction under Budget $B$}}
We evaluate how much accuracy improves when a human annotator—assumed to be always correct—uses a limited budget \(B\) to manually revise the most uncertain cases. Items are ranked by the various uncertainty signals, and the top \(B\) fraction is corrected; the resulting dataset-level accuracy after these ideal corrections is shown in Figure~\ref{fig:performance_corrected}. Each curve corresponds to a different ranking method: the proposed PE estimate (e.g., PE(Generated, MaxP+H)), the a posteriori PE/SE baseline, perplexity, random selection, and an oracle that ranks strictly by true low-accuracy severity. The gray oracle curve defines the upper bound, while the yellow random curve gives a reference for uninformed correction. \textcolor{black}{Across the evaluated models, uncertainty-guided correction substantially outperforms random selection, with the largest marginal gains at small budgets. The single‑shot regressors closely track the multi‑shot PE/SE baselines, recovering most of their improvement while reducing LLM calls from N to 1, whereas the Perplexity baseline produces consistently lower curves. TableLlama and Gemma‑2‑9B exhibit steeper initial slopes, indicating a high concentration of correctable errors among the most‑uncertain items, while other models obtain comparable gains only at larger budget fractions $B$.}

\subsection{\textcolor{black}{Q3: Learnability of the uncertainty regressor $h_\phi$}}
\label{sec:learnability}
\begin{figure}[!t]
    \centering
    \includegraphics[width=0.7\textwidth]{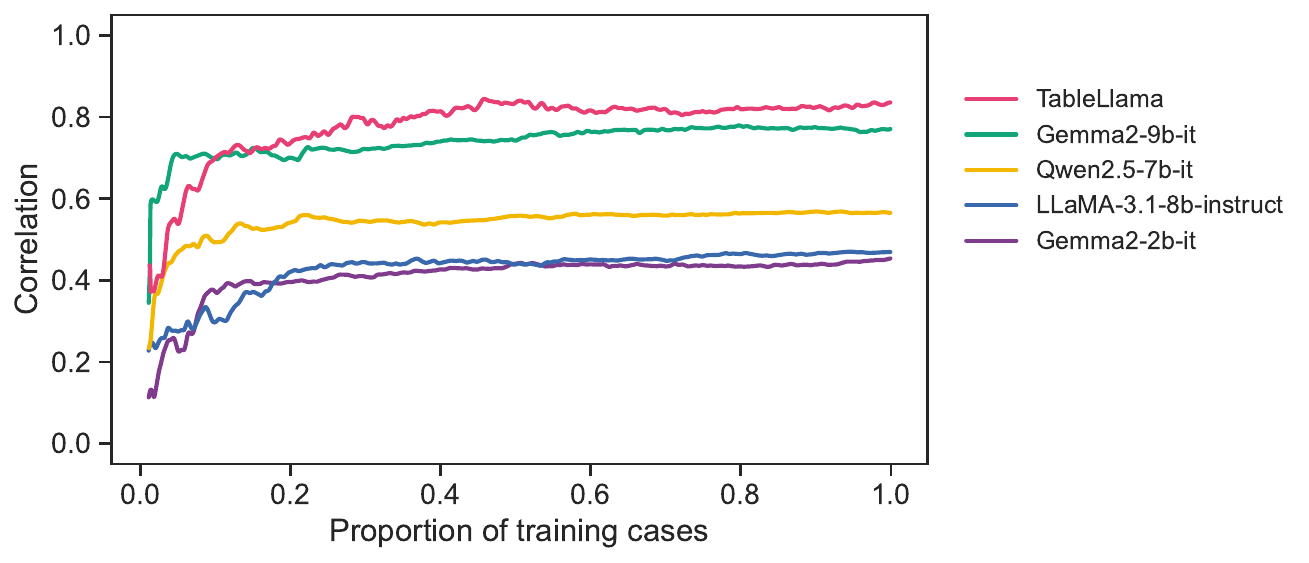}
    \caption{Spearman correlation ($\rho$) with multiple-generations PE when training the proposed method over an increasing number of training cases, average over 10-fold cross-validation.}
    \label{fig:progressive_training}
\end{figure}

Since our self-supervised regressor $h_\phi$ requires a warm-up phase for learning, we conducted two supplementary experiments to assess its practical applicability. First, we estimated the number of examples that are necessary to reach the stability of the \textcolor{black}{regressors}' performance. As a complementary analysis, we assess if a proxy target measure can be derived, trading some accuracy for a reduced computational cost.
\paragraph{\textbf{Regressor Convergence as a Function of Training Size}} We train the \textcolor{black}{regressor} with an increasing number of cases and assess its Spearman correlation with the target. We perform a 10-fold cross-validation, where in each fold we keep the validation set fixed and train the \textcolor{black}{regressor} by expanding the training set one-by-one. Figure~\ref{fig:progressive_training} reports the average performance over all the folds for each model. We regress the PE target only, using the M($\mathbf{p}$) and H$(\mathbf{p})$ features on the first $10$ \textit{Generated} tokens. Stable performance is reached with a limited number of items, e.g., $10\%-20\%$ of the dataset, depending on the model. This highlights how, even with a limited number of examples, the approach can successfully learn to estimate uncertainty. This, in turn, impacts the overall efficiency and sustainability of the method, since the number of multiple-generations cases necessary to train the model appears to be limited. We recall that, since the method is self-supervised, it does not rely on external annotations.

\begin{figure}[!t]
    \centering
    \includegraphics[width=0.7\textwidth]{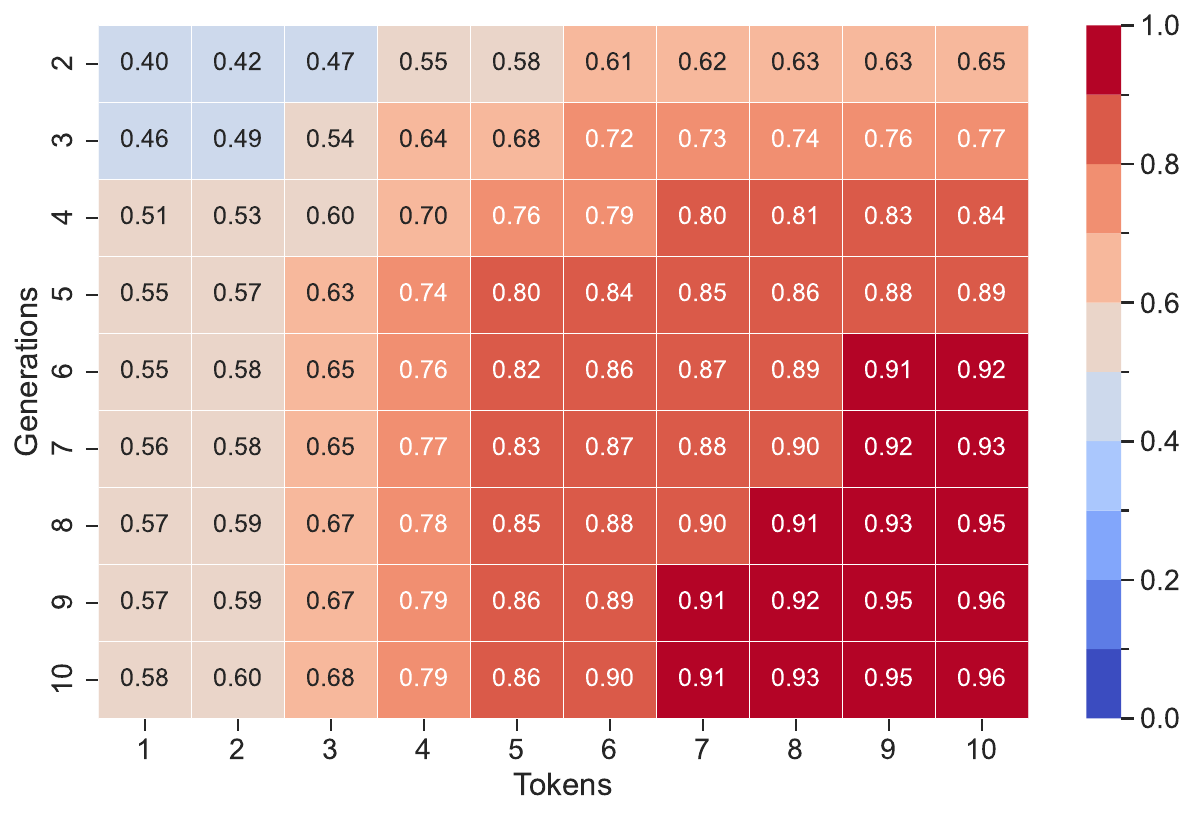}
    \caption{Spearman correlation ($\rho$) with PE, obtained with $N=10$ generations from TableLlama using all the tokens, as a function of the number of tokens (x-axis) and generations (y-axis) used to compute a truncated PE.}
    \label{fig:truncated_correlation}
\end{figure}

\paragraph{\textbf{Approximated Ground Truth}} We finally consider whether reducing the resources dedicated to the multiple generations can retain enough information to approximate the full uncertainty -- that is, the uncertainty observed with the original $N$ generations when considering all the generated tokens. For this experiment, performed for illustration purposes, we focus on the best-performing model, TableLlama, and examine the PE target only. We cap the number of generations and the number of generated tokens, both of which impact the computational burden linearly.  Figure~\ref{fig:truncated_correlation} shows the Spearman correlation between the approximated PE and the full PE. In this setup, even under computational constraints, the derived approximate PE is reasonably correlated with the full PE, i.e., computed on the complete set of generations and tokens. This highlights how an approximate target signal could be derived by trading accuracy for efficiency.

\subsection{\textcolor{black}{Q4: Temperature Sensitivity}}
\label{sec:temperature}

Our approach proposes a measure of uncertainty related to output variability. However, the temperature setting affects the trade-off between output variability -- needed to estimate uncertainty -- and task accuracy. In this section, we evaluate the impact of the temperature on the task performance. We summarize the performance as the area under the curve (AUC) that describes dataset-level accuracy as a function of correction budget $B$ (cf. Figure~\ref{fig:performance_corrected} for comparison). We chose this metric as it summarizes the practical downstream usability of the proposed method.

\begin{figure}[!t]
    \centering
    \includegraphics[width=0.8\textwidth]{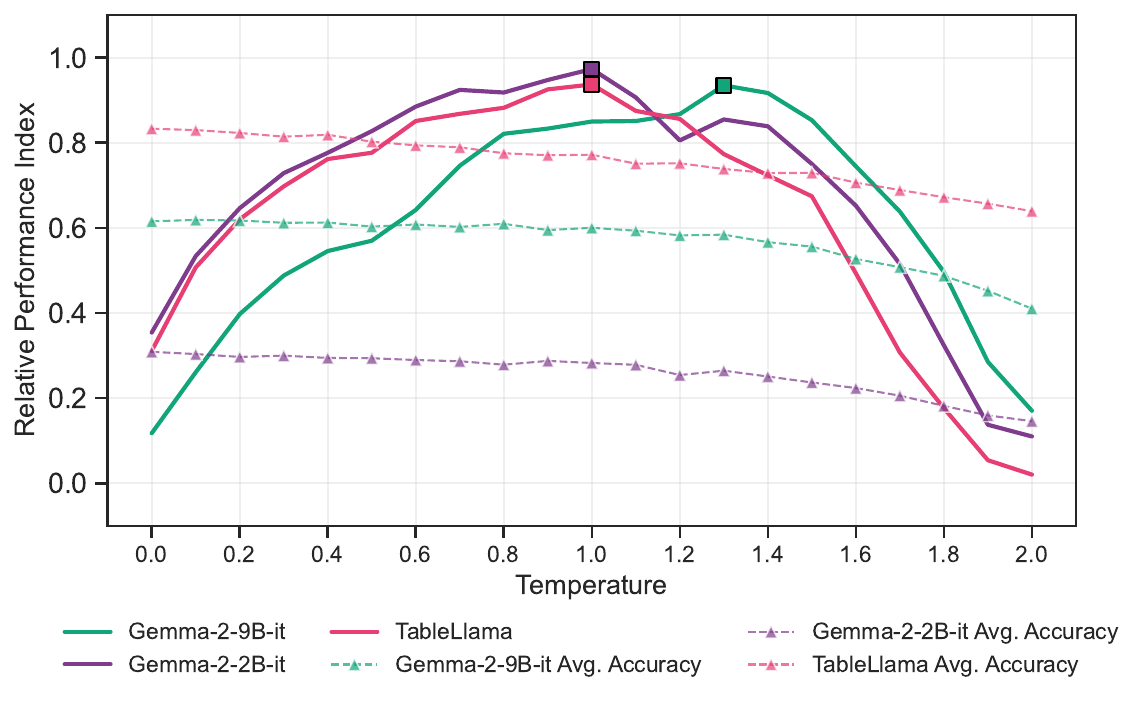}
    \caption{Sensitivity of model performance to temperature. Model performance is summarized as the area under the curve of accuracy as a function of budget (cf. Figure~\ref{fig:performance_corrected}), with selected items ranked by PE, and rescaled to [0, 1] for each model. The square markers highlight the maximum Relative Performance Index.}
    \label{fig:temperature}
\end{figure}

We systematically sweep the temperature value in the range $0.0 \leq \tau \leq 2.0$ with steps of $0.1$ to assess its effect on the AUC. Due to computational budget constraints, we utilize a subset of 200 elements from the original dataset and focus on PE. Figure~\ref{fig:temperature} reports the AUC over the temperature and the average overall accuracy obtained with selected models. It can be observed that the most convenient performance is obtained for balanced $\tau$ values, for which dataset-level accuracy does not change drastically, implying that temperature variations up to a certain level do not compromise the task performance. Setting smaller $\tau$ values degrades the performance, since lower output variability directly reduces observable uncertainty. Similarly, setting a higher $\tau$ value leads to higher observed uncertainties, which are related to noisy outputs and do not reflect the actual answer uncertainty. Based on these observations, we set $\tau=1.0$ for all the experiments previously discussed in this paper. Nonetheless, this experiment illustrates how temperature can be adjusted depending on the specific scenario.

\section{Conclusions and Future Work}
\label{sec:conclusions}

In this work, we propose a lightweight, self-supervised approach for obtaining uncertainty estimates. An approximate entropy measure is regressed by leveraging features that are observable at inference time, based on final- and intermediate-layer token probability distributions. This approximation is calculated, after an initial warm-up phase, without relying on multiple, costly generation runs, making it suitable for practical use in real-world EL workflows. 

We validate the proposed method on the task of Entity Linking for tabular data, showing a strong correlation with uncertainty observed over multiple generations estimates, with significantly reduced computational overhead. Empirical evaluations, performed across several instruction-tuned LLMs, demonstrate that the method is highly effective at identifying low-accuracy outputs, in particular when using features derived from generated tokens. We also quantitatively assess the number of cases needed for learning to predict uncertainty, showing that a limited warm-up phase is sufficient. Finally, we measure how the LLM temperature influences the trade-off between uncertainty and the overall accuracy of the method.

The proposed method is general, and it may benefit other closed-form tasks beyond the considered context. Future experiments, aimed at testing its generalizability across tasks and domains -- particularly in open-ended EL settings, where different uncertainty measures could have different behaviors and efficiency trade-offs -- could extend the results provided in this study. A warm-up phase, involving multiple LLM generations, is required for training, which introduces a computational cost. However, this cost is mitigated by the rapid convergence of the learning phase, maintaining the efficiency of the overall process. Additionally, evaluating the transferability of learned \textcolor{black}{regressors} across datasets related to the same task may provide further insights into the robustness of the method.

\textcolor{black}{While our method is model-agnostic and applies unchanged to any model exposing token log-probabilities (and optionally hidden states), we leave the profiling and evaluation of larger-capacity models to future work. When only output-layer probabilities are available (e.g., closed models), our output-layer-only variant remains applicable.}

We also show that some of the features related to the intermediate layers during the generation process are partially able to fill the performance gap observed between prompt-related and generation-related tokens. The use of internal LLM state features to anticipate properties of the generated output represents a promising research direction.
Overall, the results reinforce the practical applicability of uncertainty-aware methods in LLM-based EL workflows, particularly in contexts where quality, efficiency, and scalability are critical. Beyond mention-level scores, a natural extension is to target the specific decision tokens (e.g., the span that selects a candidate, or yes/no tokens in classification prompts) to obtain position-aware confidence; for multi-answer outputs, per-span uncertainties could further support selective review and partial acceptance.


\begin{acknowledgments}
The work has received funding from the European Union’s Horizon Europe research and innovation programme un- der grant agreements No. 101189771 (DataPACT) and No. 101070284 (enRichMyData), and the Italian PRIN project Discount Quality for Respon- sible Data Science (202248FWFS). 
\end{acknowledgments}

\section*{Declaration on Generative AI}
During the preparation of this work, the authors used ChatGPT (GPT-4-turbo) in order to perform grammar and spelling checks. After using this tool, the authors reviewed and edited the content as needed and take full responsibility for the publication’s content.

\bibliography{bibb}


\clearpage
\appendix

\newcommand{\reducedstrut}{\vrule width 0pt height .9\ht\strutbox depth .9\dp\strutbox\relax}
\newcommand{\yellowh}[1]{%
  \begingroup
  \setlength{\fboxsep}{0pt}%
  \colorbox{Yellow}{\reducedstrut#1\/}%
  \endgroup
}
\newcommand{\yellowgreenh}[1]{%
  \begingroup
  \setlength{\fboxsep}{0pt}%
  \colorbox{YellowGreen}{\reducedstrut#1\/}%
  \endgroup
}
\newcommand{\redh}[1]{%
  \begingroup
  \setlength{\fboxsep}{0pt}%
  \colorbox{RedOrange}{\reducedstrut#1\/}%
  \endgroup
}

\section{Illustrative Example}
\label{sec:illustrative}
\label{app:illustrative}

The example below illustrates the approach to resolve a mention in a table to its corresponding entity, using an LLM. A high-level \textbf{instruction}, describing the entity linking task, is provided. A \textbf{question} contains the \textbf{mention} to be linked, together with a list of entity \textbf{candidates} extracted from a retriever, e.g., Wikidata Lookup Service or LamAPI. The candidates are provided in the following format: \texttt{<label [DESC] description [TYPES] type1, type2, ..., typeK>}. A snapshot of the \textbf{input} table is provided, i.e., N rows above and below the row containing the mention, in Markdown format. The LLM selects a candidate from the predefined list and outputs it verbatim as an \textbf{answer}.

\begin{tcolorbox}[title=Entity Linking Prompt Example,colback=gray!5!white,colframe=gray!80!black,fonttitle=\bfseries]
\small
\textbf{Instruction:} \textit{This is an entity linking task. The goal for this task is to link the selected entity mention in the table cells to the entity in the knowledge base. You will be given a list of referent entities, with each one composed of an entity name, its description and its type. Please choose the correct one from the referent entity candidates.} [...]

\vspace{0.5em}
\textbf{Input:} \texttt{[TLE] List of high schools in South Dakota.} \\
\texttt{col: |school|type|city|county|mascot|} \\
\texttt{row 0: |Aberdeen High School|Private|Aberdeen|Brown|Knights|} \\
\texttt{row 1: |Agar High School|Public|Agar|Sully|} \\
\texttt{[...]}\\
\texttt{row 35: |Crazy Horse High School|Public|Wanblee|Jackson|Chiefs|} \\
\texttt{row 36: |Crow Creek High School|Public|Stephan|\yellowh{Hyde}|Chieftains|} \\ 
\texttt{row 37: |Custer High School|Public|Custer|Custer|Wildcats|} \\
\texttt{[...]}\\

\vspace{0.5em}
\textbf{Question:} \textit{The selected entity mention is: `Hyde'. The column name for `Hyde' is `county'. The referent candidates are:} \\
\texttt{<Dr.~Jekyll and Mr.~Hyde [DESC] fictional characters [TYPE] group of fictional characters>} \\
\texttt{<Hyde v Hyde [DESC] landmark case of the English Court of Probate and Divorce [TYPE] legal case>} \\
\texttt{<Hyde County [DESC] county in South Dakota, United States [TYPE] county of South Dakota>} \\
\texttt{<Douglas Hyde [DESC] first President of Ireland (1860-1949) [TYPE] linguist>} \\
\texttt{<Hyde County [DESC] county in North Carolina, United States [TYPE] county of North Carolina>} \\
\texttt{<Hyde [DESC] civil parish in Bedfordshire, UK [TYPE] civil parish>} \\
\texttt{<Hyde Park [DESC] town in Dutchess County, New York, United States [TYPE] town of New York>} \\
\texttt{<Strange Case of Dr Jekyll and Mr Hyde [DESC] novel by R. L. Stevenson [TYPE] literary work>} \\
\texttt{<Hyde Park [DESC] neighborhood in Chicago, Illinois [TYPE] neighborhood>} \\
\texttt{[...]}\\

\textit{What is the correct referent entity for `Hyde'?}

\vspace{0.5em}
\textbf{Postilla:} \textit{Answer with just a candidate, selected from the provided referent entity candidates list, and nothing else. The selected candidate must be reported verbatim from the list provided as input. Each candidate in the list is enclosed between < and > and reports [DESC] and [TYPE] information.}

\vspace{0.5em}
\textbf{Answer:}\mbox{ }\texttt{<Hyde County [DESC] county in South Dakota, United States [TYPE] county of South Dakota>}
\end{tcolorbox}

\section{Representative Qualitative Examples}

We provide representative examples of the strengths and shortcomings of our method, reporting for each example: input table, mention, candidate list, generated answers with their frequencies, baseline, and estimated uncertainties. \textbf{Recoverable error 1} reports the standard case: a model's output is uncertain, with borderline accuracy (0.5), and our one-shot prediction successfully highlights such uncertainty. \textbf{Recoverable error 2} reports a more complex case, in which the same mention is referred to over multiple seasons. As a consequence, the model struggles to fix on one single entity. Again, our method is able to recover this case by predicting an uncertain outcome. We also report an unrecoverable error case, in which there is no answer variability. Our method correctly predicts low answer variability and, consequently, the case is not marked for correction. However, the answer is wrong, and the error cannot be recovered. Notably, the regressor accurately captures uncertainty across all the examples.
\vspace{1em}

\begin{tcolorbox}[title={Recoverable error 1: high answer variability $\rightarrow$ low accuracy},colback=gray!5!white,colframe=gray!80!black,fonttitle=\bfseries]
\small

\begin{center}
\begin{tabular}{p{4.4cm}|p{2cm}|l|l}
\hline
\texttt{party} & \texttt{city} & \texttt{province} & \texttt{registration} \\
\hline
\texttt{Democracia Galega} & \texttt{Oleiros} & \texttt{A Coruña} & \texttt{1997-01-13} \\
\texttt{Partido Democrático Español} & \cellcolor{Yellow}\texttt{Madrid} & \texttt{Madrid} & \texttt{1997-01-13} \\
\texttt{Partido Nacional Republicano} & \texttt{Valladolid} & \texttt{Valladolid} & \texttt{1997-01-13} \\
\texttt{Partido del Amor Universal} & \texttt{Barcelona} & \texttt{Barcelona} & \texttt{1997-02-12} \\
\texttt{Els Verds de Formentera} & \texttt{Formentera} & \texttt{Balearic Islands} & \texttt{1997-02-20} \\
\texttt{Els Verds d'Eivissa} & \texttt{Ibiza} & \texttt{Balearic Islands} & \texttt{1997-02-20} \\
\texttt{}  & \texttt{...} & \texttt{} & \texttt{} \\
\end{tabular}

\vspace{0.5ex}
Table: Registered political parties in Spain, \yellowh{mention} highlighted
\end{center}


\begin{center}
\begin{tabular}{p{3.6cm}|p{3.6cm}|p{4cm}}
\hline
\texttt{Name} & \texttt{Description} & \texttt{Type} \\
\hline
\cellcolor{RedOrange}\texttt{Madrid} & \cellcolor{RedOrange}\texttt{Spanish Congress Electoral District} & \cellcolor{RedOrange}\texttt{electoral district of the Spanish Congress} \\
\texttt{Madrid} & \texttt{city in Iowa, United States} & \texttt{city in the United States} \\
\texttt{Madrid Province} & \texttt{province of Spain (1833–)} & \texttt{province of Spain} \\
\texttt{Madrid} & \texttt{constituency of the Senate of Spain} & \texttt{constituency of the Senate of Spain} \\
\texttt{Madrid} & \texttt{poem by Alfred de Musset} & \texttt{version, edition, or translation} \\
\cellcolor{YellowGreen}\texttt{Madrid} & \cellcolor{YellowGreen}\texttt{capital city of Spain} & \cellcolor{YellowGreen}\texttt{municipality of Spain} \\
\texttt{Madrid} & \texttt{None} & \texttt{passenger ship} \\
\cellcolor{RedOrange}\texttt{Madrid} & \cellcolor{RedOrange}\texttt{None} & \cellcolor{RedOrange}\texttt{electoral district} \\
\texttt{Madrid} & \texttt{film} & \texttt{film} \\
\texttt{Madrid} & \texttt{mountain in South Africa} & \texttt{mountain} \\
\texttt{Madrid} & \texttt{encyclopedia article} & \texttt{encyclopedia article} \\
\texttt{} & \centering\texttt{...} & \texttt{}
\end{tabular}

\vspace{0.5ex}
Candidate entities with description and type (\yellowgreenh{right} and \redh{wrong} answers)
\end{center}

\begin{flushleft}
\hspace{1em}Observed answers (count):\\
\begin{tabular}{@{}p{12cm} r@{}}
\hspace{2em}\texttt{<Madrid [DESC] capital city of Spain [TYPE] municipality of Spain>} & \texttt{\textbf{\colorbox{YellowGreen}{(5)}}} \\
\hspace{2em}\texttt{<Madrid [DESC] Spanish Congress Electoral District } & \textbf{} \\
\hspace{2em}\texttt{[TYPE] electoral district of the Spanish Congress>} & \texttt{\textbf{\colorbox{RedOrange}{(3)}}} \\
\hspace{2em}\texttt{<Madrid [DESC] None [TYPE] electoral district>} & \texttt{\textbf{\colorbox{RedOrange}{(2)}}} \\
\end{tabular}

\bigskip

\hspace{1em}\texttt{
\makebox[4cm][l]{Predictive Entropy:}%
\makebox[1cm][l]{0.649}%
\makebox[5.2cm][l]{Predicted using our method (avg):}%
\colorbox{Yellow}{0.588}
}

\hspace{1em}\texttt{
\makebox[4cm][l]{Semantic Entropy:}%
\makebox[1cm][l]{0.478}%
\makebox[5.2cm][l]{Predicted using our method (avg):}%
\colorbox{Yellow}{0.421}
}
\end{flushleft}

\end{tcolorbox}

\begin{tcolorbox}[title={Recoverable error 2: high answer variability $\rightarrow$ low accuracy},colback=gray!5!white,colframe=gray!80!black,fonttitle=\bfseries]
\small

\begin{center}
\begin{tabular}{p{1.5cm}|p{2cm}|l|l|l}
\hline
\texttt{Season} & \texttt{Club} & \texttt{League} & \texttt{Apps} & \texttt{Goals} \\
\hline
\texttt{1986–87} & \texttt{Fiorentina} & \cellcolor{Yellow}\texttt{Serie A} & \texttt{5} & \texttt{1} \\
\texttt{1987–88} & \texttt{Fiorentina} & \cellcolor{Yellow}\texttt{Serie A} & \texttt{27} & \texttt{6} \\
\texttt{} & \texttt{} & \texttt{...} & \texttt{} & \texttt{} \\
\texttt{1994–95} & \texttt{Juventus} & \cellcolor{Yellow}\texttt{Serie A} & \texttt{17} & \texttt{8} \\
\texttt{1995–96} & \texttt{Milan} & \cellcolor{Yellow}\texttt{Serie A} & \texttt{28} & \texttt{7} \\
\texttt{} & \texttt{} & \texttt{...} & \texttt{} & \texttt{} \\
\texttt{1998–99} & \texttt{Inter} & \cellcolor{Yellow}\texttt{Serie A} & \texttt{23} & \texttt{6} \\
\texttt{1999–00} & \texttt{Inter} & \cellcolor{Yellow}\texttt{Serie A} & \texttt{18} & \texttt{6} \\
\texttt{} & \texttt{} & \texttt{...} & \texttt{} & \texttt{} \\
\end{tabular}

\vspace{0.5ex}
Table: Roberto Baggio career stats, \yellowh{mention} highlighted
\end{center}


\begin{center}
\begin{tabular}{p{3.6cm}|p{3.6cm}|p{3.6cm}}
\hline
\texttt{Name} & \texttt{Description} & \texttt{Type} \\
\hline
\cellcolor{YellowGreen}\texttt{Serie A} & \cellcolor{YellowGreen}\texttt{top Italian football league} & \cellcolor{YellowGreen}\texttt{annual sporting event} \\
\texttt{Serie A (basketball) 2003–04} & \texttt{sports season} & \texttt{sports season} \\
\texttt{Serie A (basketball) 2007–08} & \texttt{None} & \texttt{sports season} \\
\texttt{Serie A} & \texttt{2nd tier of Italian women's rugby union} & \texttt{national championship} \\
\texttt{Serie A} & \texttt{top Italian pallapugno league} & \texttt{sports competition} \\


\texttt{1984–85 Serie A} & \texttt{sports season} & \texttt{sports season} \\
\cellcolor{RedOrange}\texttt{1986–87 Serie A} & \cellcolor{RedOrange}\texttt{sports season} & \cellcolor{RedOrange}\texttt{sports season} \\
\texttt{1990–91 Serie A} & \texttt{sports season} & \texttt{sports season} \\
\cellcolor{RedOrange}\texttt{1994–95 Serie A} & \cellcolor{RedOrange}\texttt{sports season} & \cellcolor{RedOrange}\texttt{sports season} \\
\texttt{1997–98 Serie A} & \texttt{sports season} & \texttt{sports season} \\
\cellcolor{RedOrange}\texttt{1998–99 Serie A} & \cellcolor{RedOrange}\texttt{sports season} & \cellcolor{RedOrange}\texttt{sports season} \\
\texttt{1999–2000 Serie A} & \texttt{sports season} & \texttt{sports season} \\
\texttt{} & \centering\texttt{...} & \texttt{}
\end{tabular}

\vspace{0.5ex}

Candidate entities with description and type (\yellowgreenh{right} and \redh{wrong} answers)
\end{center}

\begin{flushleft}
\hspace{1em}Observed answers (count):\\

\begin{tabular}{@{}p{12cm} r@{}}
\hspace{2em}\texttt{<1986--87 Serie A [DESC] sports season [TYPE] sports season>} & \texttt{\textbf{\colorbox{RedOrange}{(4)}}} \\
\hspace{2em}\texttt{<1994--95 Serie A [DESC] sports season [TYPE] sports season>} & \texttt{\textbf{\colorbox{RedOrange}{(3)}}} \\
\hspace{2em}\texttt{<Serie A [DESC] top Italian football league [TYPE] annual sporting event>} & \texttt{\textbf{\colorbox{YellowGreen}{(2)}}} \\
\hspace{2em}\texttt{<1998--99 Serie A [DESC] sports season [TYPE] sports season>} & \texttt{\textbf{\colorbox{RedOrange}{(1)}}} \\
\end{tabular}

\bigskip

\hspace{1em}\texttt{
\makebox[4cm][l]{Predictive Entropy:}%
\makebox[1cm][l]{0.639}%
\makebox[5.2cm][l]{Predicted using our method (avg):}%
\colorbox{Yellow}{0.571}
}

\hspace{1em}\texttt{
\makebox[4cm][l]{Semantic Entropy:}%
\makebox[1cm][l]{0.638}%
\makebox[5.2cm][l]{Predicted using our method (avg):}%
\colorbox{Yellow}{0.559}
}
\end{flushleft}

\end{tcolorbox}


\begin{tcolorbox}[title={Unrecoverable error: no answer variability \& zero accuracy},colback=gray!5!white,colframe=gray!80!black,fonttitle=\bfseries]
\small

\begin{center}
\small
\begin{tabular}{p{1.5cm}|l|p{1.cm}|l|p{3cm}|m{1cm}}
\hline
\texttt{name} & \texttt{family} & \texttt{language} & \texttt{region} & \texttt{country} & \texttt{pop (M)} \\
\hline
\texttt{Agaw} & \texttt{Cushitic} & \texttt{Agaw} & \texttt{Horn of Africa} & \texttt{Ethiopia. Eritrea} & \texttt{1} \\
\texttt{Amhara} & \texttt{Semitic} & \texttt{Amharic} & \texttt{Horn of Africa} & \texttt{Ethiopia} & \texttt{20} \\
\texttt{Beja} & \texttt{Cushitic} & \texttt{Beja} & \texttt{Horn of Africa} & \texttt{Sudan. Eritrea} & \texttt{2} \\
\texttt{Bilen} & \texttt{Cushitic} & \cellcolor{Yellow}\texttt{Bilen} & \texttt{Horn of Africa} & \texttt{Eritrea} & \texttt{0.2} \\
\texttt{Gurage} & \texttt{Semitic} & \texttt{Gurage} & \texttt{Horn of Africa} & \texttt{Ethiopia} & \texttt{1.9} \\
\texttt{Oromo} & \texttt{Cushitic} & \texttt{Afan Oromo} & \texttt{Horn of Africa} & \texttt{Ethiopia. Somalia. Sudan. Kenya} & \texttt{30} \\
\texttt{Saho} & \texttt{Cushitic} & \texttt{Saho} & \texttt{Horn of Africa} & \texttt{Eritrea. Ethiopia} & \texttt{0.2} \\
\texttt{} & \texttt{} & \centering \texttt{...} & \texttt{} & \texttt{} & \texttt{} \\ 
\end{tabular}

\vspace{0.5ex}

Table: Ethnic groups in Horn of Africa, \yellowh{mention} highlighted
\end{center}

\begin{center}
\small
\begin{tabular}{p{3.6cm}|p{3.6cm}|p{3.6cm}}
\hline
\texttt{Name} & \texttt{Description} & \texttt{Type} \\
\hline
\texttt{Bilen i dag} & \texttt{Swedish periodical} & \texttt{periodical} \\
\texttt{Blågula Bilen} & \texttt{aid agency} & \texttt{aid agency} \\
\texttt{Bilen \& miljön} & \texttt{Swedish periodical} & \texttt{periodical} \\
\texttt{Handsfreetelefoni i bilen} & \texttt{motion by Inger Jarl Beck et al. 2005} & \texttt{individual motion} \\
\cellcolor{RedOrange}\texttt{Bilen people} & \cellcolor{RedOrange}\texttt{ethnic group} & \cellcolor{RedOrange}\texttt{ethnic group} \\
\texttt{Bilen} & \texttt{1992 film by John Goodwin} & \texttt{film} \\
\texttt{Ensam i bilen} & \texttt{article in Drömmen om bilen (1997)} & \texttt{article} \\
\cellcolor{YellowGreen} \texttt{Blin} & \cellcolor{YellowGreen} \texttt{language} & \cellcolor{YellowGreen} \texttt{modern language} \\
\texttt{När bilen drabbade landsbygden} & \texttt{article in Drömmen om bilen (1997)} & \texttt{article} \\
\texttt{Bilen} & \texttt{family name} & \texttt{family name} \\
\texttt{Mobilförbud i bilen} & \texttt{motion by Helena Bargholtz 2009} & \texttt{individual motion} \\
\texttt{} & \centering \texttt{...} & \texttt{} \\ 

\end{tabular}

\vspace{0.5ex}

Candidate entities with description and type (\yellowgreenh{right} and \redh{wrong} answers)
\end{center}

\begin{flushleft}
\hspace{1em}Observed answers (count):\\

\begin{tabular}{@{}p{12cm} r@{}}
\hspace{2em}\texttt{<Bilen people [DESC] ethnic group [TYPE] ethnic group>} & \texttt{\textbf{\colorbox{RedOrange}{(10)}}} \\
\end{tabular}

\bigskip

\hspace{1em}\texttt{
\makebox[4cm][l]{Predictive Entropy:}%
\makebox[1cm][l]{0.0}%
\makebox[5.2cm][l]{Predicted using our method (avg):}%
\colorbox{Yellow}{0.044}
}

\hspace{1em}\texttt{
\makebox[4cm][l]{Semantic Entropy:}%
\makebox[1cm][l]{0.0}%
\makebox[5.2cm][l]{Predicted using our method (avg):}%
\colorbox{Yellow}{0.041}
}
\end{flushleft}

\end{tcolorbox}

\clearpage
\section{Position-Dependent Feature Contributions}
\label{app:position-dep-contrib}
\begin{figure}[!h]
    \centering
    \includegraphics[width=0.8\textwidth]{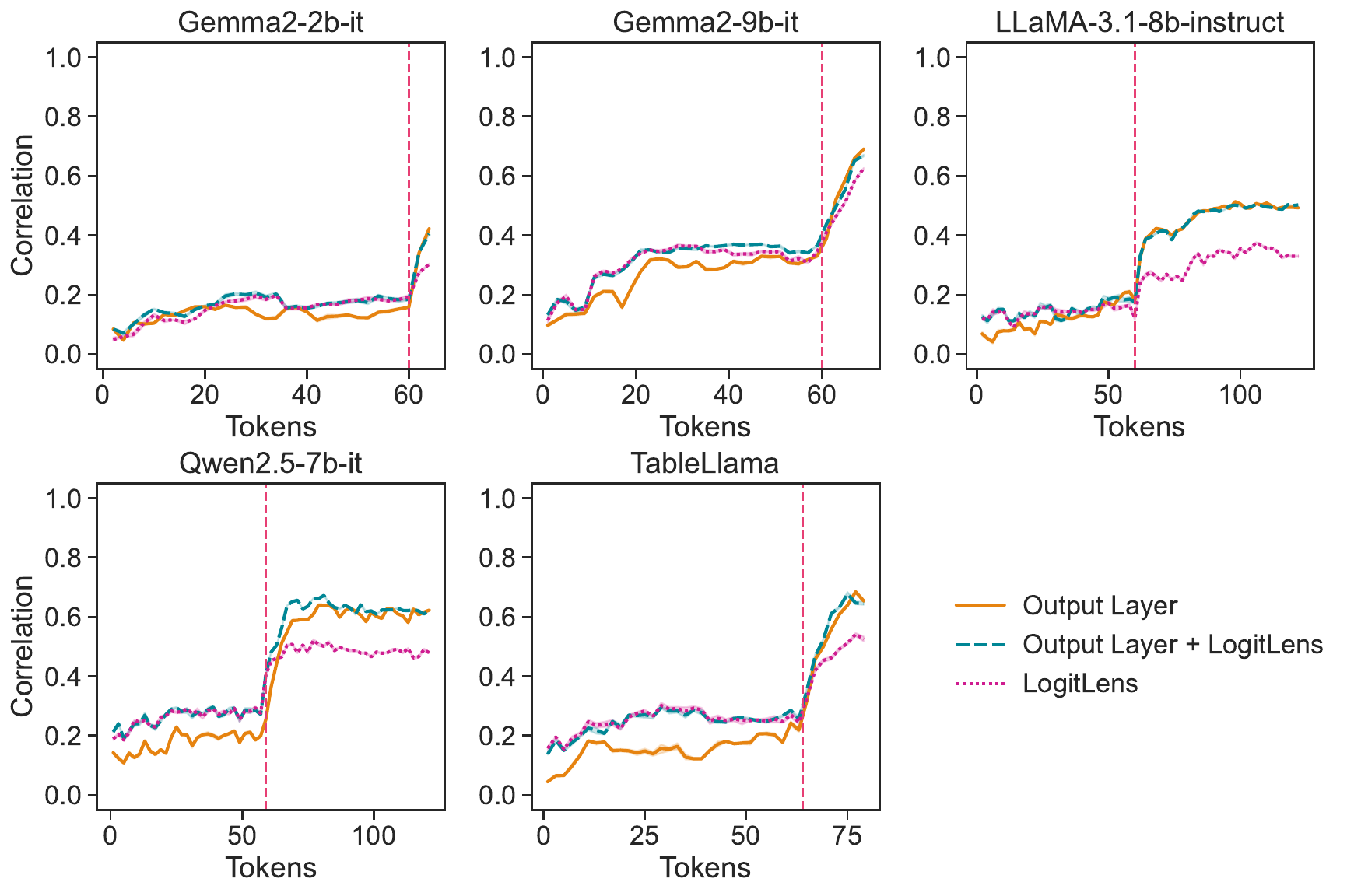}
    \caption{Spearman correlation with PE when training on a progressive window of tokens spanning the concatenated \textit{Postilla} and \textit{Generated} segments. The vertical dashed line marks the boundary between the two segments.}
    \label{fig:growing_window}
\end{figure}

\textcolor{black}{Figure~\ref{fig:performance} reveals that performance differs depending on whether features are extracted from the \textit{Postilla} or \textit{Generated} tokens. To investigate this further, we designed an experiment that progressively expands a sliding window over the concatenated \textit{Postilla} and \textit{Generated} segments, measuring how information accumulates as more tokens are included. We evaluate three feature configurations: (1) output-layer features, (2) LogitLens features from intermediate layers, and (3) their combination. Figure~\ref{fig:growing_window} shows the Spearman correlation between each configuration and the baseline PE as a function of the window size. Correlation increases gradually while the window traverses the \textit{Postilla} tokens, with a pronounced jump once the \textit{Generated} tokens are reached. The improvement over \textit{Postilla} is non-uniform: certain positions—especially the final tokens—contribute disproportionately, indicating that the informative signal is unevenly distributed. Additionally, LogitLens features provide a benefit within the \textit{Postilla} segment but not after entering the \textit{Generated} portion, suggesting that during generation, the output-layer features already capture sufficient information, whereas intermediate-layer representations are more useful prior to generation. These observations imply that feature selection should be adapted based on token origin and position to maximize effectiveness. \textcolor{black}{This also motivates extensions that localize uncertainty at the answer-bearing token(s), enabling per-token confidences and finer-grained handling of multi-answer outputs.}}





\clearpage
\section{Spearman Correlation Between Estimated and True PE/SE}

\begin{figure}[htbp]
    \centering
    \includegraphics[width=0.88\textwidth]{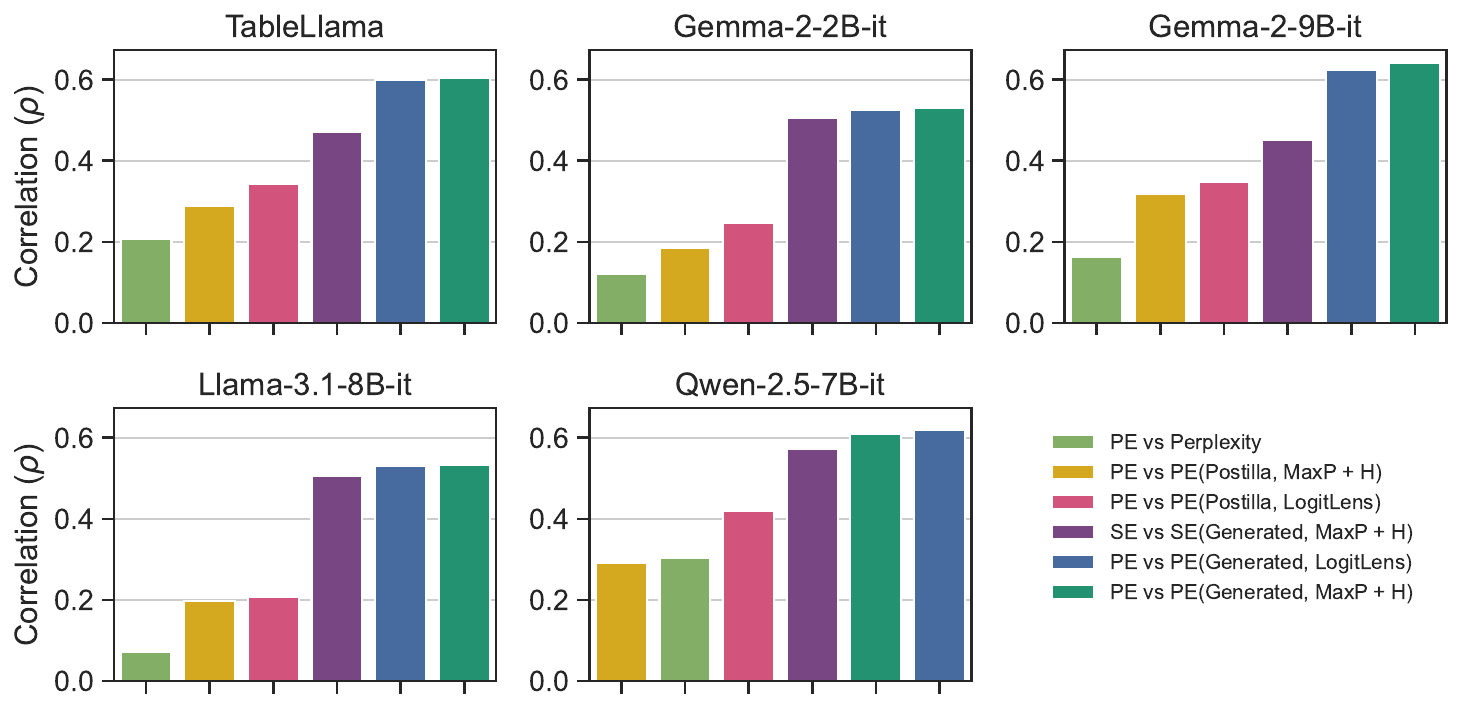}
    \caption{Spearman's rank correlation coefficient ($\rho$) between the estimated entropy (PE or SE) and the multiple-generations entropy (PE or SE) across models. In the legend, the notation ``\texttt{Target(Segment, Observable)}'' indicates that the target variable \textit{Target} was predicted using a regressor based on \textit{Observable} features extracted from the \textit{Segment} segment of the prompt.}
    \label{fig:spearman-corr}
\end{figure}
\FloatBarrier

\section{Answer Variability as a Function of Temperature}

\begin{figure}[htbp]
    \centering
    \includegraphics[width=0.95\textwidth]{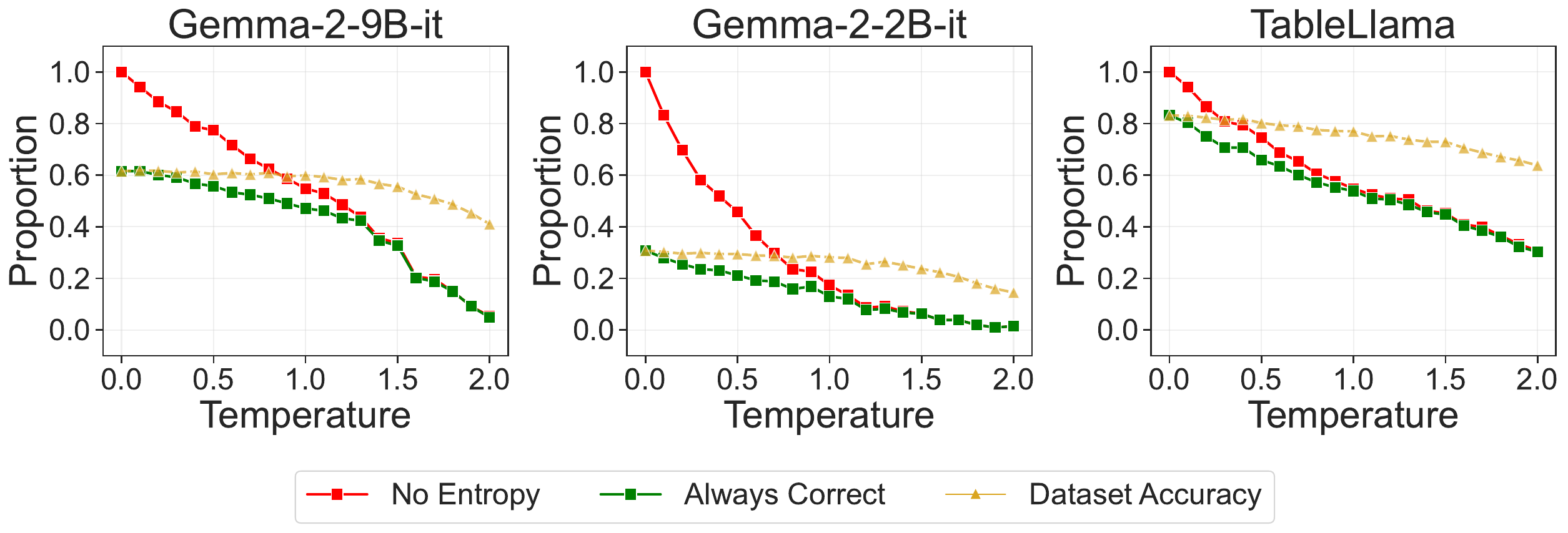}
    \caption{Sensitivity to temperature of prompts with no output variation (\textit{red}), prompts with correct answers only -- still no output variation (\textit{green}), and prompts with correct answer on average ($\geq 0.5$, \textit{yellow}),  \textbf{(left)} Gemma-2-9b-it, \textbf{(center)} Gemma-2-2b-it, and \textbf{(right)} TableLlama.}
    \label{fig:temperature_old}
\end{figure}
\FloatBarrier

To evaluate the variability of answers and recoverable cases, we systematically sweep temperature values in the range $0.0 \leq \tau \leq 2.0$ with steps of $0.1$ to assess the temperature effect on (1) the number of items that show output variability and (2) the average accuracy. For capacity reasons, we utilize a subset of 200 elements from the original dataset.
Figure~\ref{fig:temperature_old} shows the results of the experiment for two selected LLMs. The \textit{yellow} series shows average accuracy, highlighting how higher $\tau$ values deteriorate the performance on the task. At the same time, lower $\tau$ settings yield the highest proportion of always correct cases (\textit{green} line). However, again at low $\tau$, output variability is also lowest; the uncertainty of these cases cannot be estimated, making this setting unsuitable for recovery. Operationally, a reasonable trade-off for $\tau$ is to minimize the accuracy loss while maximizing the number of cases recoverable through uncertainty, that is, those above the \textit{red} line. Another way to view this trade-off is by examining the difference between the \textit{red} and \textit{green} lines: this difference represents the proportion of unrecoverable cases, that is, cases that are always wrong and have no output variability.  Making these cases recoverable requires ``paying'' by reducing always correct cases, while trying to preserve average accuracy (\textit{yellow}) as much as possible. This experiment illustrates how temperature can be adjusted depending on specific combinations of model, task, and application constraints.

\section{Transformer-Based Time Complexity}
\label{app:complexity}

To further assess the practicality and usability of our approach, in this section we derive the time complexity of a Transformer-based architecture, such as the ones used by the models considered in this work. We consider a Transformer with $L$ layers, and $d$ hidden size. The context length is set to $N$, while the number of generated tokens is $G$. The time complexity of a single forward pass over a prompt $X\in\mathbb{R}^{N \times d}$can be decomposed into the following components:

\begin{itemize}
    \item \textbf{Self-attention:} Given the $Q,K,V \in \mathbb{R}^{N \times d}$ matrices, the time complexity of the self-attention mechanism is $O(N^2 \cdot d)$, where $N$ is the context length and $d$ is the hidden size. The attention scores are computed as $QK^T$, and the output is computed as $\sigma(QK^T)V$. The time complexity of this operation is $O(N^2 \cdot d)$.
    \item \textbf{Feed-forward:} During the feed-forward step, the time complexity is $O(8N \cdot d^2)=O(N \cdot d^2)$ overall, where $d$ is the hidden size. In this we can include also the projection of the input to the $QKV$ space and the final projection to the output space.
\end{itemize}

\noindent In total one has a time complexity of $O(N^2 \cdot d + N \cdot d^2)$ for a single Transformer layer, which becomes $O(L[N^2 \cdot d + N \cdot d^2])$, where $L$ is the number of layers.

If we now suppose to generate $G$ tokens without the use of a KV-cache, the time complexity of the $g$-th generation step is $O(L[(N+g)^2 \cdot d + (N+g) \cdot d^2])$, for every $g \in [G]$. If we then sum over all the generations, we have:

\begin{align}
O\left(\sum_{g=1}^{G}L\left[(N+g)^2 d + (N+g) d^2\right]\right)&=\\
O\left(L \sum_{g=1}^{G} \left[(N+g)^2 d + (N+g) d^2 \right] \right)&=\\
O\left(L \sum_{g=1}^{G} \left[(N^2 + 2Ng + g^2)d + (N+g) d^2 \right] \right)&=\\
O\left(L \left[ (G N^2 + 2NG^2 + G^3)d + (GN+G^2)d^2 \right] \right)&=\\
O\left(L \left[ G(N+G)^2d + G(N+G)d^2 \right]  \right)&=\\
O\left(LG \left[(N+G)^2d + (N+G)d^2 \right]  \right)
\end{align}

which shows that the time complexity of generating $G$ tokens is quadratic in the number of overall tokens $N+G$, when $G \ll N$, otherwise it would become cubic in the number of generated ones.

When a KV-cache is used, while the time for processing the prompt is the same, a major computational saving is obtained during the generation phase. In this case, the time complexity of the $g$-th generation step can be decomposed into the following components:

\begin{itemize}
    \item \textbf{Self-attention:} During the self-attention, $Q$ reduces to a single vector $q_g \in \mathbb{R}^{1 \times d}$, while $K,V$ becomes $K_g, V_g \in \mathbb{R}^{(N+g-1) \times d}$. Overall, the time complexity of this operation is $O((N+g) \cdot d)$.
    \item \textbf{Feed-forward:} Since the output of the Self-Attention has a dimension of $1 \times d$, the feed-forward step has a time complexity of this operation is $O(d^2)$, which includes the projection of the input to the $QKV$ space and the final projection to the output space.
\end{itemize}

\noindent If we then sum over all the generations $g \in [G]$, we have:

\begin{align}
O\left(\sum_{g=1}^{G}L\left[(N+g) d + d^2\right]\right)&=\\
O\left(L \sum_{g=1}^{G} \left[(N+g) d + d^2 \right] \right)&=\\
O\left(L \left[ G(N+G)d + Gd^2 \right] \right)&=\\
O\left(LG \left[ (N+G)d + d^2 \right] \right)&
\end{align}

\noindent which shows that the time complexity of generating $G$ tokens is linear in the number of overall tokens $N+G$, when $G \ll N$, otherwise it would become quadratic in the number of generated ones.

\begin{table}[!h]
\centering
\setlength{\tabcolsep}{9pt}
\caption{Time complexity of the Transformer-based architecture. $L$ is the number of layers, $N$ is the number of tokens in the prompt, $G$ is the number of generated tokens, and $d$ is the hidden size.}
\vspace{0.5ex}
\label{tab:complexity}
\begin{tabular}{c c c} 
 \toprule
 \textbf{Phase} & \textbf{Without KV-cache} & \textbf{With KV-cache} \\ [0.25ex] 
 \midrule
 Prompt processing & $O(L[N^2 \cdot d + N \cdot d^2])$ & $O(L[N^2 \cdot d + N \cdot d^2])$ \\
 Generation & $O(LG[(N+G)^2d + (N+G)d^2])$ & $O(LG[(N+G)d + d^2])$ \\ [0.25ex]
 \bottomrule    
\end{tabular}
\end{table}

This simple derivation shows that our proposed approach to learn to estimate the multiple-generations Predictive Entropy (PE) is still computationally promising even when advanced KV-cache techniques are employed during inference.
\end{document}